\documentclass[twoside,11pt]{article}

\usepackage[abbrvbib, preprint]{jmlr2e}

\usepackage{booktabs}
\usepackage{amsmath}
\usepackage{algorithm2e}

\jmlrheading{1}{2022}{1-48}{11/21}{__/__}{____}{Divya Bhargavi, Erika Pelaez Coyotl and Sia Gholami}

\ShortHeadings{Identifying football player jersey numbers with synthetic data}{Bhargavi, Pelaez Coyotl and Gholami}
\firstpageno{1}

\begin{document}

\title{Knock, knock. Who's there?--Identifying football player jersey numbers with synthetic data}

\author{\name Divya Bhargavi \email dbharga@amazon.com \\
  \name Erika Pelaez Coyotl \email erpelaez@amazon.com \\
  \name Sia Gholami \email gholami@amazon.com \\
  \addr Amazon Web Services, CA USA}

\editor{__}

\maketitle

\begin{abstract}%

  Automatic player identification is an essential and complex task in sports video analysis. Different strategies have
  been devised over the years, but identification based on jersey numbers is one of the most common approaches given its
  versatility and relative simplicity. However, automatic detection of jersey numbers is still challenging due to
  changing camera angles, low video resolution, small object size in wide-range shots and transient changes in the
  player's posture and movement. In this paper we present a novel approach for jersey number identification in a small,
  highly imbalanced dataset from the Seattle Seahawks practice videos. We use a multi-step strategy that enforces
  attention to a particular region of interest (player's torso), to identify jersey numbers. We generate in-house
  synthetic datasets of different complexities to supplement the data imbalance and scarcity in the samples. Our
  multi-step pipeline first identifies and crops players in a frame using a pretrained person detection model.
  We then utilize a pretrained human pose estimation model to localize jersey numbers (using torso key-points) in
  the detected players, obviating the need for annotating bounding boxes for number detection. This results in images
  that are on average 20x25px in size. We trained two light-weight Convolutional Neural Networks (CNNs) with different
  learning objectives: multi-class for two-digit number identification and multi-label for digit-wise detection to
  compare performance. Both models went through a pre-training round with the synthetic datasets and were finetuned
  with the real-world dataset to achieve a final best accuracy of 89\%. Our results indicate that simple models can
  achieve an acceptable performance on the jersey number detection task and that synthetic data can improve the
  performance dramatically (accuracy increase  of ~9\% overall, ~18\% on low frequency numbers) making our approach
  achieve state of the art results.

\end{abstract}

%
%

\section{Introduction}\label{sec:introduction}

\begin{figure}[h]
  \centering
  \includegraphics[width=\linewidth]{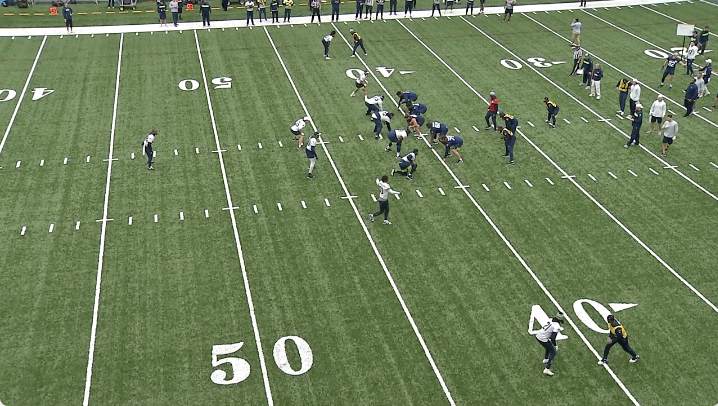}
  \caption{Example frames from the practice videos demonstrating the challenges to identify jersey numbers in zoomed out videos.}\label{fig:wideshot}
\end{figure}

\begin{figure}[h]
  \centering
  \includegraphics[width=\linewidth]{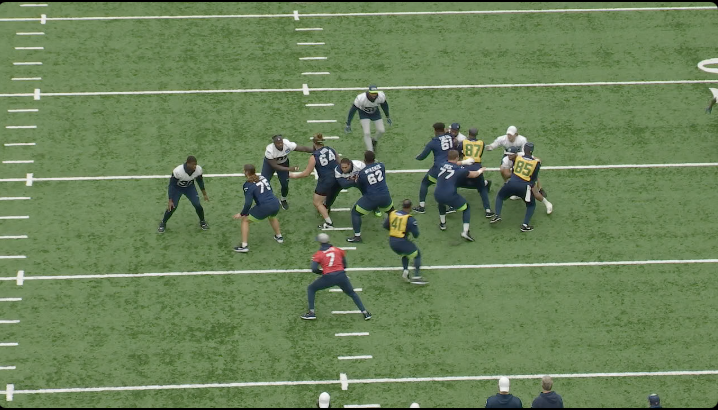}
  \caption{Cropped players examples showing the player posture, movement and camera angle challenges to
  identify jersey numbers.}\label{fig:playerposture}
\end{figure}

In recent years, interest in analyzing team sport videos has increased significantly in academia and
industry~\citep{r1, r2, r3, r4, r5, r6, r7}.
This is important for sports broadcasters and teams to understand key events in the game and
extract useful information from the videos. Use cases include identifying participating players, tracking player movement
for game statistics, measuring health and safety indicators, and automatically placing graphic overlays.
For broadcasters and teams that don't have the leeway or the capital to install hardware sensors in player wearables,
a Computer Vision (CV) based solution is the only viable option to automatically understand and generate insights
from games or practice videos. One important task in all sports CV applications is identifying players, specifically
identifying players with their jersey numbers. This task is challenging due to distortion and deformation of player
jerseys based on the player posture, movement and camera angle, rarity of labelled datasets, low-quality videos,
small image size in zoomed out videos, and warped display caused by the player movement.
(see Figure~\ref{fig:wideshot} and ~\ref{fig:playerposture}) \par

Current approaches for jersey number identification consist of two steps: collecting and annotating large
datasets~\citep{r5, r7}, and training large and complex models~\citep{r5, r6, r7}. These approaches include either sequential training of
multiple computer vision models or training one large model, solving for 2 objectives: identifying the jersey number
location (through custom object detection models or training a custom human pose estimation model) and classifying
the jersey number~\citep{r4, r5, r6, r7}. These approaches are tedious, time-consuming, and cost-prohibitive thus making it
intractable for all sports organizations. \par

In this paper we present a novel approach to detect jersey numbers in a small dataset consisting of practice video
footage from the Seattle Seahawks team . We use a three-step approach to number detection that leverages pretrained
models and novel synthetic datasets. We first identify and crop players in a video frame using a person detection model.
We then utilize a human pose estimation model for localizing jerseys on the detected players using the torso key-points,
obviating the need for annotating bounding boxes for number locations. This results in images that are less than
20x25 px with a high imbalance in jersey numbers (see Figure~\ref{fig:playerposture}). Finally, we test two different learning approaches
for model training - multi-class and multi-label each yielding an accuracy of 88\%, with an ensemble accuracy of
89\% to identify jersey numbers from cropped player torsos. \par

Additionally, to compensate for the low number of examples in some of the jersey numbers, we propose two novel
synthetic dataset generators — Simple2D and Complex2D. The Simple2D generator creates two-digit number images from
different combinations of fonts and background colors to mimic those of the Seattle Seahawks jerseys. The Complex2D
generator superimposes the Simple2D numbers on random COCO dataset~\citep{r8} images to add more complexity to the background
and make the model training robust. By pretraining our two CNNs on these synthetic datasets, we observe a 9\% increase
in accuracy on the ensemble models pre-trained with synthetic data compared to the baseline models trained with the
only the Seattle Seahawks numbers. Furthermore, we observe better generalization with low data. \par

\section{Related work}\label{sec:related-work}

\subsection{Synthetic Data Generation}\label{subsec:rw-synthetic-data-generation}
CNN algorithms, that are commonly used in most CV tasks, require large datasets to learn patterns in images.
Collecting and annotating large datasets is a manual, costly and time-consuming task. Several new approaches
including Active Learning~\citep{r9}, Zero or Few-shot learning~\citep{r10} and Synthetic data generation~\citep{r11} have emerged in
recent years to tackle complexities in obtaining a large annotated dataset. Our work focuses primarily on the use
of synthetically generated data. This idea dates back to the 1990's~\citep{r12} and is an active field of research that
alleviates the cost and efforts needed to obtain and manually label real-world data. Nowadays, models (pre)trained
on synthetic datasets have a broad range of utility including feature matching~\citep{r13} autonomous driving~\citep{r14}, robotics
indoor and aerial navigation~\citep{r15}, scene segmentation~\citep{r16} and anonymized image generation in healthcare~\citep{r17}.
The approaches broadly adopt the following process: pre-train with synthetic data before training on real-world
scenes~\citep{r13, r18}, generate composites of synthetic data and real images to create a new one that contains the desired
representation~\citep{r19} or generate realistic datasets using simulation engines like Unity~\citep{r20} or generative models
like GANs~\citep{r21, r22}. There are limitations to each of these regimes but one of the most common pitfalls is
performance deterioration in real-world datasets. Models trained only synthetic datasets don't generalize to
real-world data; this phenomenon is called "domain shift"~\citep{r21}.
\par
In order to reduce the need for annotating large dataset as well as account for the size and imbalance of the
real-world data, we generated two double-digit synthetic datasets - Simple2D and Complex2D with different levels
of complexity as described in Section~\ref{subsubsec:syn-data-gen} This helps to circumvent the domain shift when only synthetic data is
used and improves generalization on real-world data for fine-tuning.

\subsection{Number Identification}\label{subsec:rw-number-identification}
Automatic number identification in sports video has evolved from classical computer vision techniques including
feature extraction using contrast adjustment, edge detection of numbers~\citep{r1, r2, r3} to deep learning-based architectures
that use CNNs for classification~\citep{r4, r5, r6, r7}. A fundamental problem in number identification in sports is the
jersey number distortion due to erratic and continuous player movement. The spatial transformer-based approach
introduced in~\citep{r5} tries to localize and better position the number, so that the classifier has a better chance of
an accurate prediction. The faster-RCNN with pose estimation guidance mechanism~\citep{r6} combines the detection,
classification and key-point estimation tasks in one large network to correct region proposals, reducing the
number of false negative predictions. This approach needed careful labeling of the player bounding-boxes and four
human body key-points, shoulder (right, left), hip (right, left), in addition to the numbers. It also made use of
high-resolution number images (512 px). This approach yields 92\% accuracy for jersey number recognition as a whole
and 94\% on the digit-wise number recognition task. However, getting the right conditions for it i.e., label the
dataset for the three tasks, acquiring high resolution images and training a large model might be challenging for
real-world cases. Furthermore, a lack of standardization and availability of public (commercial use) datasets,
makes it difficult to obtain a benchmark for the number identification task.

\section{Approach}\label{sec:approach}

\subsection{Task Definition}\label{subsec:task-definition}
We define a jersey number as the one or two-digit number printed on the back of a player’s shirt. The jersey number is
used to identify and distinguish players and one number is associated with exactly one player. Our solution takes
cropped images of player’s torsos as input and attempts to classify the jersey number into 101 classes
(0-99 for actual numbers and 100 for unrecognizable images/ jerseys with no numbers).

\subsection{American Football Dataset}\label{subsec:american-football-dataset}
\begin{figure}[h]
  \centering
  \includegraphics[width=\linewidth]{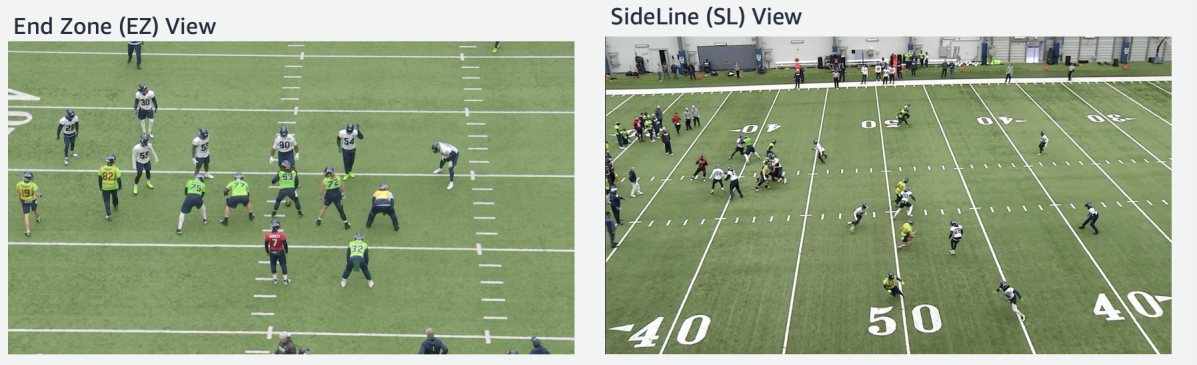}
  \caption{Examples of frames obtained from the two different angles from the training videos. Left, is the endzone
  view of the players. Right is the sideline view which offers better visibility into jersey numbers. Within a play,
    we can find players, observers with/without football jerseys.}\label{fig:perspectives}
\end{figure}

The data used for this work consisted of a collection of 6 practice videos from different angles for training and
additional 4 for testing from the Seattle Seahawks archives. Half of the videos were from the endzone perspective,
that is, the scoring zone between the end line and the goal line. The other half were from the sideline perspective,
the boundary line that separates the play area from the sides. Both cameras were placed on a high altitude to get a
panoramic view for the play and capture the majority of the actions taken by the players. A pitfall for collecting
data using this camera angle is that the size of a player is less than 10\% of the image size when the players are
far away from the camera. In addition, the sideline view has restricted visibility of jersey numbers compared to
end-zone (see Figure~\ref{fig:perspectives}). The videos were recorded in 1280x720 resolution and we sampled frames
from each video at 1, 5 and 10 frames per second (fps) rates. We noticed that images sampled at 5 fps sufficiently
captured all the jersey numbers in a play and we decided to use the same sampling rate throughout our solution.

\subsubsection{Jersey number localization}\label{subsec:jersey-number-localization}
To mitigate the need for annotating player location, jersey number bounding boxes and consequently training person and
jersey number detection models, we utilized pretrained models for person detection and pose estimation to localize the
jersey number region. This approach prevents the model to generate correlations with wrong features like player
background, helmets or clothing items and confining the learning to the region of interest.

\begin{figure}[h]
  \centering
  \includegraphics[width=\linewidth]{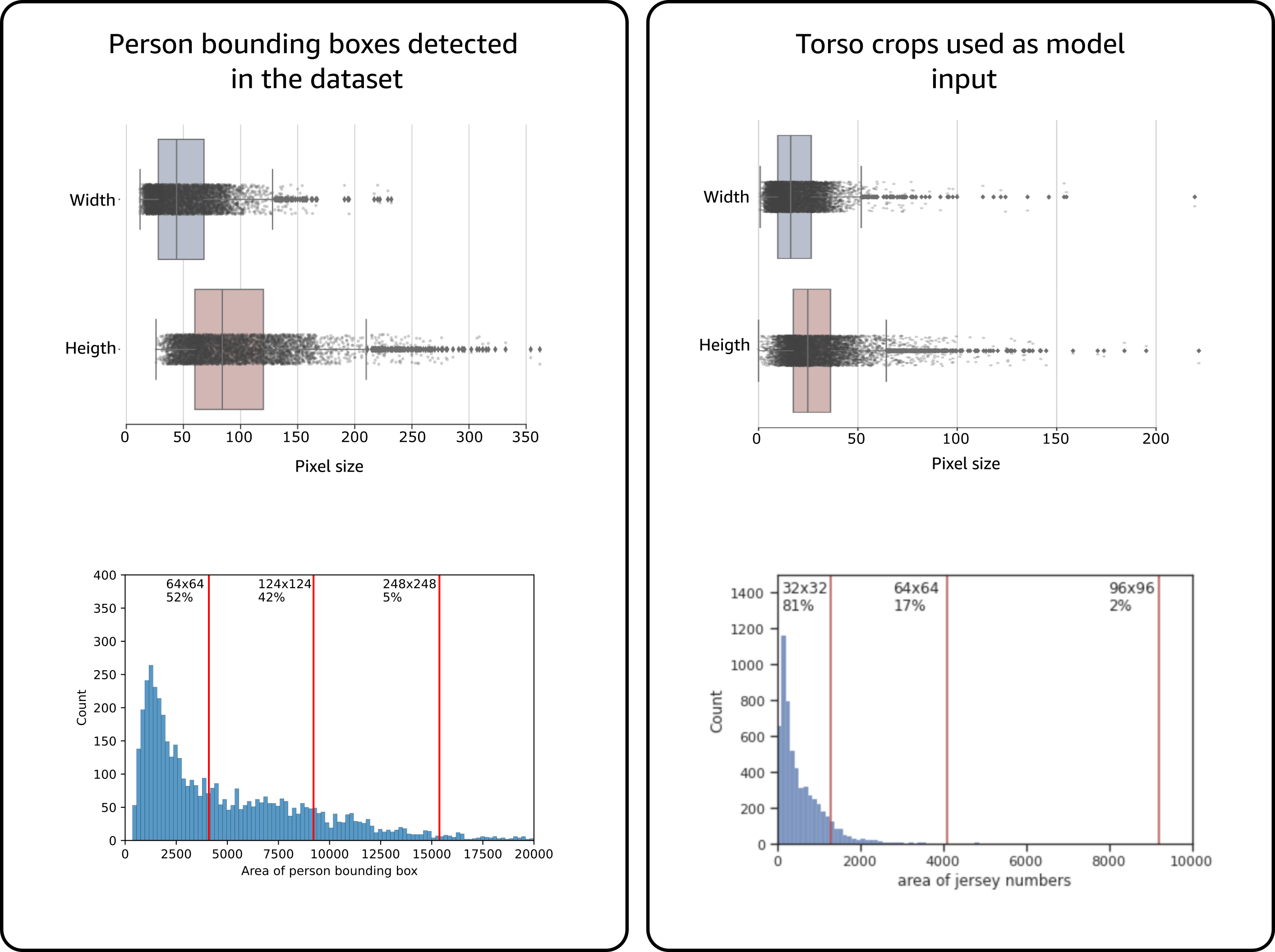}
  \caption{Distribution of the sizes from person and torso bounding boxes. Note how the great majority of torso sizes is less than 32x32 px.}\label{fig:datasize}
\end{figure}

For the number localization we first use a pretrained person detector, Centernet~\citep{r23} model (ResNet50 backbone), to
detect and crop players from an image. Instead of training a custom human key-point estimation head~\citep{r6}, we use a
pretrained, pose estimation model, AlphaPose (https://gitee.com/marcy/AlphaPose, with ResNet101 backbone), to identify
four torso key-points
(left and right - hips and shoulders) on the cropped player images from the person detection step (see Figure~\ref{fig:models}).
We use the four key-points to create a bounding box around jersey numbers. To accommodate inaccuracies in key-point
prediction and localization due to complex human poses, we increased the size of torso keypoint area by expanding the
coordinates 60\% outward to better capture jersey numbers. The torso area is then cropped and used as the input for
the number prediction models discussed in Section~\ref{subsubsec:syn-data-gen} In previous works, the use of high-resolution images of
players and jersey numbers is very common. However, the American football dataset we used was captured from a bird’s
eye view, where jersey numbers were smaller than 32x32 px. In fact, the average size of the torso crops is 20x25 with
the actual jersey number being even a smaller portion of this area (see Figure~\ref{fig:datasize}).

\begin{figure}[h]
  \centering
  \includegraphics[width=\linewidth]{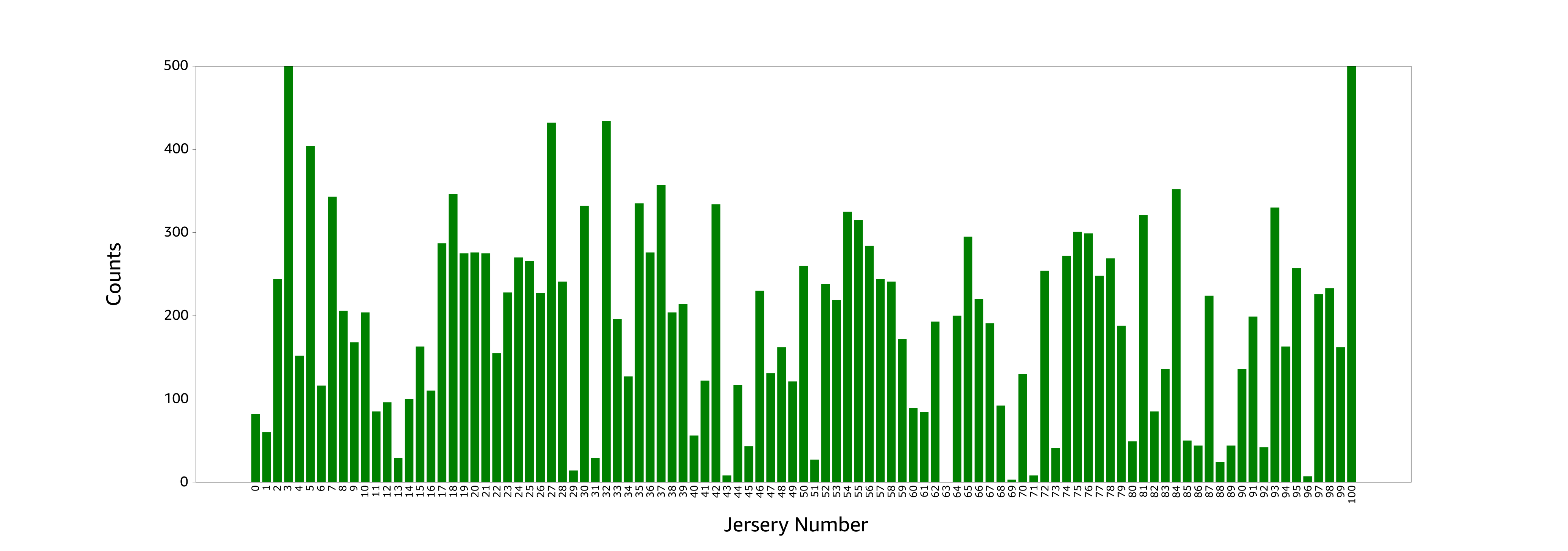}
  \caption{Distribution of the jersey number labels in training set. Number 3 has 500+ images while numbers 43, 63, 69 and 93 have 10 images or less.}\label{fig:datadistro}
\end{figure}

After player detection and jersey number localization, we generated 9,000 candidate images for number detection.
We labelled the images with Amazon SageMaker GroundTruth  and noticed that 6,000 images contained non-players
(trainers, referees, watchers); the pose estimation model for jersey number localization simply identifies human
body key-points and doesn’t differentiate between players and non-players. 3,000 labelled images with severe
imbalance (see Figure~\ref{fig:datadistro}) were usable for the training.

\subsubsection{Synthetic Data Generation}\label{subsubsec:syn-data-gen}
Typically, a licensed (SVHN~\citep{r25}) or a large custom dataset is used for (pre)training number recognition models.
Since there are no standardized public datasets with permissive licenses, we created two 2-digit synthetic datasets
to pretrain our models. We investigated 2-digit MNIST~\citep{r26}, however it did not have pixel color and font variations
needed for jersey detection and performed poorly in our tests. Hence, we generated two different synthetic datasets;
a simple two-digit (Simple2D) numbers with font and background similar to the football dataset and other with 2-digit
synthetic numbers superimposed on COCO~\citep{r8} dataset images (Complex2D) to account for variations in numbers background.

The Simple2D dataset was generated by randomly selecting a number from a uniform distribution of 0 to 9 and randomly
scaling it. Color backgrounds (Red, Navy Blue, Green, Red, Yellow, White) and special font (Freshman ) that resembled
the team jerseys were used to generate these numbers (see Figure~\ref{fig:datasize}). One Light, five Medium and five Hard augmentations
(see Table~\ref{tab:data-aug}) were used on each digit to be later permuted and concatenated to obtain 4000 images (100 x 100 px) of
each 2-digit number, from 00 to 99. At the end this dataset consisted of a total of 400,000 images.

\begin{figure}[h]
  \centering
  \includegraphics[width=\linewidth]{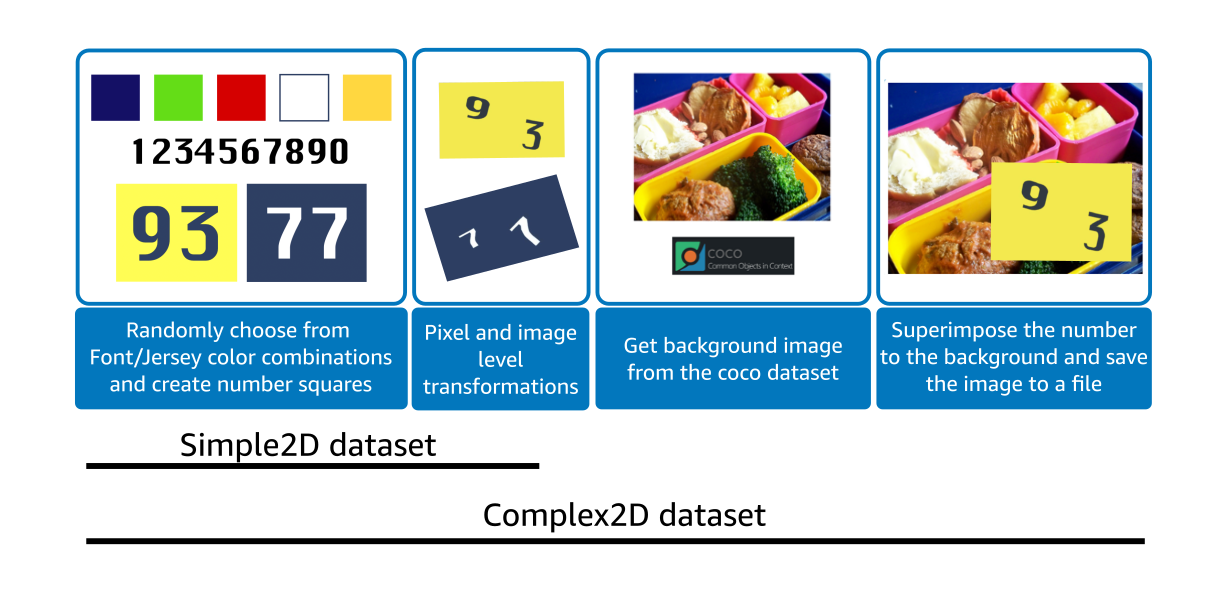}
  \caption{Synthetic data generation with Simple2D and Complex2D. Simple2D dataset was generated by creating numbers
  in football dataset jersey colors and fonts. Several augmentations (Table~\ref{tab:data-aug}) were applied on these numbers to get
  Simple2D dataset. The numbers from this dataset were randomly sampled and randomly placed on COCO dataset images
  to form Complex2D dataset}\label{fig:synthetic}
\end{figure}

Since the real-world images had more complicated background, textures and lighting conditions, we decided to
synthetically generate another dataset (see Figure~\ref{fig:synthetic}) to increase the robustness and generalization of our pretrained
model. The complex2D dataset was designed to increase background noise by superimposing numbers from Sample2D on
random real-world images from the COCO dataset~\citep{r8}. We generated a total of 400,000 images (4000 per class) with
noisy backgrounds.
Our algorithm is explained in more details in Algorithms~\ref{alg:number-generation}, \ref{alg:simple2d} and \ref{alg:complex2d}.

\begin{table}[h]
  \caption{data augmentations}\label{tab:data-aug}
  \centering
  \begin{tabular}{p{.1\linewidth} p{0.85\linewidth}}
    \toprule
    Name & Augmentations    \\
    \midrule
    Light & Gaussian Noise, Optical distortion      \\
    Medium & Light + Grid distortion    \\
    Hard & Medium + Shuffling RGB channels, Random Shift-Scale-Rotation \\
    \bottomrule
  \end{tabular}
\end{table}

\begin{algorithm}[hbt!]
\caption{Number generation}\label{alg:number-generation}
\ForAll{n in 0-9}{
  select a jersey background and font color with a probability of U(1,n) = number of combinations\;
  choose a font size with a probability of U(a,b) if a, b are scaled factors of image size \;
  paste single number with chosen font and background color and size \;
}
\end{algorithm}

\begin{algorithm}[hbt!]
\caption{Simple2D}\label{alg:simple2d}
\ForAll{n in 0-99}{
  \ForAll{background colors}{
    generate 1000 images\;
    \eIf{single digit}{
      perform light, medium and hard augmentations\;
      scale image to 100x100 px\;
    }{
      perform light, medium and hard augmentations on each digit\;
      concatinate digits \;
      scale image to 100x100 px\;
    }
  }
}
randomly sample 4000 images per number across all color combinations \;
\end{algorithm}

\begin{algorithm}[hbt!]
\caption{Complex2D}\label{alg:complex2d}
\ForAll{n in 0-99}{
  select a random image from COCO dataset\;
  select a random jersey number image\;
  super-impose jersey number at a random position in the COCO image\;
  rescale image to 100x100 px\;
  continue until 4000 images per number are obtained\;
}
\end{algorithm}

\subsubsection{Jersey number detection}\label{subsubsec:jersey-n-detection}
\begin{figure}[h]
  \centering
  \includegraphics[width=\linewidth]{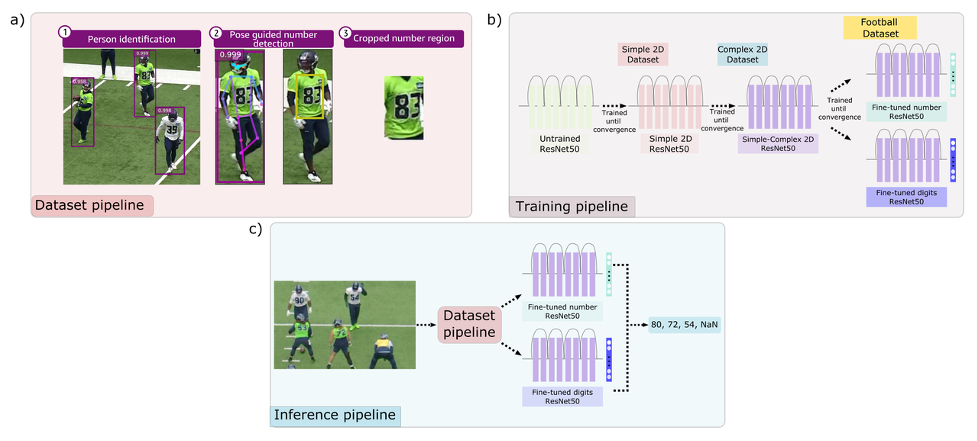}
  \caption{Overview of the approach for extracting data, training and generating jersey number predictions.
  a) describes the high-level football dataset processing pipeline - identify person in video, pass each person image
  through pose estimation model to identify torso region and crop them. b) shows the sequential pretraining of
  multi-class/label models with synthetic number datasets - Simple2D and Complex2D as well as fine-tuning on football
  dataset. c) represents the inference pipeline that uses data pipeline from a) to crop jersey numbers and perform
  prediction using multi-class/label models Figure b)}\label{fig:models}
\end{figure}

After the number localization step above, two models were sequentially pretrained with the synthetic datasets
(Simple2D to Complex2D) and fine-tuned with the real-world football dataset (see Figure~\ref{fig:models}). The idea of training a
model with increasingly difficult samples is called curriculum learning. This technique has empirically shown accuracy
increase and faster convergence~\citep{r27, r28}. One of the challenges of implementing curriculum learning is manually ranking
difficulty in the training set~\citep{r27}. In our case, the synthetic data was generated explicitly in this manner
(simple to complex) and our training regime adopted this order, thus, bypassing this challenge.

Both models used a ResNet50~\citep{r29} architecture with deep residual connections, as backbone and a final layer predicting
classes (jersey numbers). The first model was a multi-class image classifier to detect two-digit number with a total
of 101 different classes (numbers from 0 - 99 plus an unrecognizable class). The second model was a multi-class
multi-label classifier with 21 classes to detect single digits (10 digits for each side- right, left numbers, plus an
unrecognizable class).

We define the i-th input feature $X_i$ (cropped image of a player) with the label $y_i$ (0-99 for actual numbers and 100 for
unrecognizable). Our multi-class model was optimized with the following loss function:

\[ L_{mc} = \sum_{i} {L_i} = - \sum_{i} {y_i \log \hat{y}_{mc} (X_i)} \]

where $y_i$ is the true label and $\hat{y}_{mc}$ is calculated as a softmax over scores computed by the multi-class
model as follows:

\[ \hat{y}_{mc} (X_i) = \sigma (\vec{Z}) \]
\[ \sigma (\vec{Z})_k = \frac {e^{Z_k}} {\sum_{j=0}^{100} e^{Z_j}} \]

Where $\vec{Z}$ is the outputs from the last layer of the multiclass model consists of $(z_0, ..., z_100)$ given $X_i$.

For the multi-label model, the loss function is defined as:

\[ L_{ml} = \sum_{i} {L_i} = - \sum_{i} {y_i \log \hat{y}_{ml} (X_i)} \]

where $y_i$  is the true label and $\hat{y}_{ml}$ is calculated as a sigmoid over scores computed by the multi-label model as follows:
\[ \hat{y}_{ml} (X_i) = \frac {1} {1 + e^{\vec{Z}}} \]

Where $\vec{Z}$ is the outputs from the last layer of the multilabel model given $X_i$.

Both models were trained until convergence and the model from the epoch with the best performance was selected. We
explored the combination of the two models to provide the final decision and we explain our results in section~\ref{sec:exp-results}
Our original idea was that the multi-label model would augment performance of the multi-class model and address
generalization issues with unseen/ low data availability for certain numbers. For example, if 83, 74 were present in
the training set but not 73, the right and left side of prediction nodes for 3 and 7 would have been activated in the
train set for all numbers starting and ending with 7 or 3 and hence the multi-label model would have enough samples
to predict 73.

We considered training a custom object detection model to identify single-digit numbers. However, due to additional
cost and time associated with labeling bounding boxes, image quality and small size of localized jersey numbers
(approximately 20 x 25 px), we chose the image classification approach.

\section{Experimental Results}\label{sec:exp-results}
We trained the ResNet50 multi-class(number-detection) and multi-label(digit-detection) jersey number classifiers on
the football dataset to establish baseline performance without the synthetic data. For the multi-class model, we
took the number with highest softmax score as the prediction. For the multi-label model, we applied a threshold of
0.5 to both right and left predicted classes to get the output. Eventually we computed the final prediction from the
output of the two models.

The baseline model accuracy was 80\% for both models. We experimented with various input image sizes and found optimal
accuracy at 224x224 px for the multi-class and 100x100 px for the multi-label model. Our dataset presented a high
imbalance across several numbers where 24\% of the numbers have less than 100 samples and only 5\% reach the 400-sample
mark (See Figure~\ref{fig:perspectives}). Hence, we duplicated data points for each number to have 400 images in the training set when
needed. Our training pipeline dynamically applies image augmentation so that no image is seen twice by the models,
even when the base image is the same. We also up sample our test-set images to maintain 20 images per number.

After having our baselines, we investigated the effects of pre-training with the generated synthetic data on our model
performance. Pre-training on the Simple2D dataset and fine-tuning on the football dataset, resulted in a performance
improvement of 2\% over the baseline (82\%), for both, multi-class and multi-label models. However, pre-training on
the Complex2D dataset and fine-tuning on the football dataset, resulted in 3\% improvement on the multi-class model
and 8\% on the multi-label model. By pre-training on both Simple2D and Complex2D, we achieved 8.8\% and 6\% improvement
above the baseline in multi-class and multi-label models respectively.

The best multi-label model (Complex2D + Football dataset) had positive accuracy improvements on 74 classes, no change
in accuracy in 19 classes, negative change in accuracy in 8 classes (drop by 10\%).  The best multi-class model
(Simple2D + Complex2D + Football dataset) had positive accuracy improvements on 63 classes, no change in accuracy in
21 classes, negative change in accuracy in 17 classes (drop by 7\%). In order to validate the hypothesis
(Section~\ref{subsubsec:jersey-n-detection}) that multi-label model could have better performance on numbers with
less images, we compare its
results with best multi-class model on numbers with less than 50 images in training set. We notice an average increase
in accuracy of 18.5\% for multi-class model and 20\% for multi-label model before and after training on synthetic data,
for these numbers.  Despite larger gains in accuracy shown by multi-label model, the absolute accuracy scores for these
numbers were better for multi-class model, 81\% compared to 78\% for multi-label model.

\begin{figure}
  \centering
  \includegraphics[width=.35\textwidth]{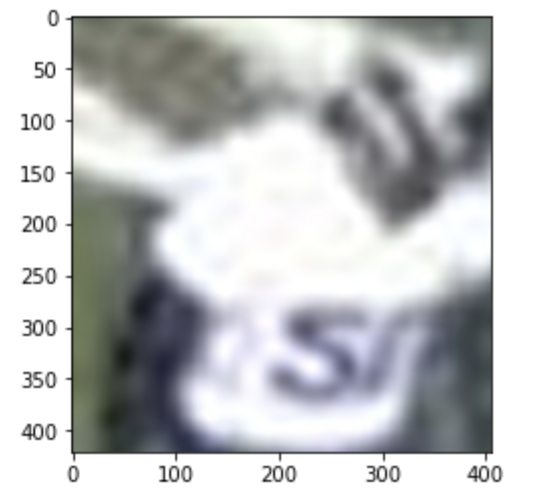}
  \includegraphics[width=.3\textwidth]{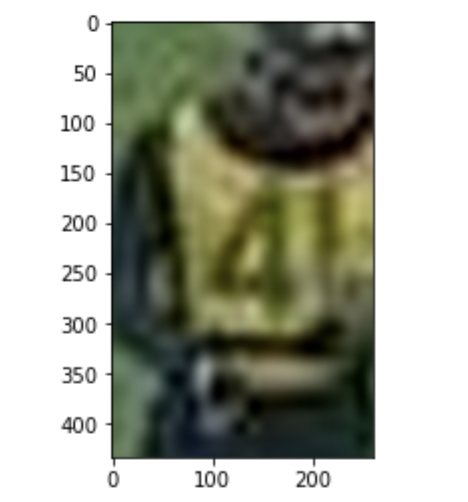}
  \caption{Images where multi-label predicted class 100. The multi-label model is not sure of the number class when
  the input image has very low resolution.}\label{fig:mc100}
\end{figure}

\begin{figure}
  \centering
  \includegraphics[width=.3\textwidth]{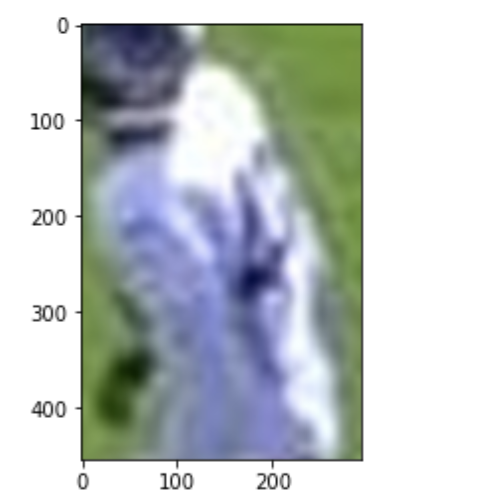}
  \includegraphics[width=.3\textwidth]{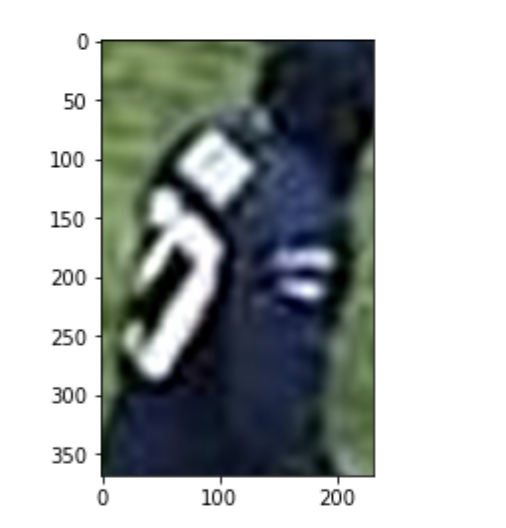}
  \caption{Image where multi-class predicted class 100. Confusion for the multi-class model arise when the
  numbers are rotated or occluded.}\label{fig:ml100}
\end{figure}

By analyzing the confusion matrix of the model predictions 
, we learnt that the best multi-label
model produces false predictions in 2 major scenarios (see Figure~\ref{fig:mc100}): predicting one digit rather than both digits,
and predicting class 100 for low-resolution and hard-to-recognize digits. In other words, the multi-label model is
more likely to predict one digit number and non-number classes when challenged with new data.  The multi-class model,
however, has relatively spread-out false predictions (see Figure~\ref{fig:ml100}). Major areas of error for this model are:
predicting one digit rather than both digits, and mistaking single digits for two digits or unrecognizable class.

Examining the performance of the two models independently we noticed that predictions agree in 84.4\% of the test
cases, suggesting that despite the different objectives (multi-class vs multi-label) there is a robust learning of
the number representations. Furthermore, we notice an additional improvement of 0.4\% by two-model ensemble.
Table 2 presents our results.

\begin{table}[h]
  \caption{A comparison of model performance under different conditions with confidence threshold of 0.5}\label{tab:results}
  \centering
  \begin{tabular}{p{.5\linewidth} p{0.1\linewidth}p{0.1\linewidth}p{0.1\linewidth}}
    \toprule
    Experiment & Multi-class & Multi-label & Ensemble    \\
    \midrule
    \multicolumn{4}{c}{Without synthetic data} \\
    Football dataset & 0.8064 & 0.8 &  \\
    Best (Multi-class + Multi-label) & & & 0.8028  \\
    \\
    \multicolumn{4}{c}{With synthetic data pre-training} \\
    Simple2D + Football dataset & 0.8282 & 0.82 & \\
    Complex2D + Football dataset & 0.8306 & 0.88 & \\
    Simple2D + Complex2D + Football dataset & 0.8886 & 0.86 & \\
    Best (Multi-class + Multi-label) & & & 0.8931  \\
    \bottomrule
  \end{tabular}
\end{table}

\section{Limitations}\label{sec:limitations}
The work presented in this paper shows that the number identification task can be simplified by leveraging synthetic
datasets. We were able to obtain a good performance that is comparable with previous works~\citep{r1, r2, r4} requiring no
change in the data collection pipeline. Despite these findings, we recognize this approach has some limitations which
we describe in this section.

We were able to achieve 89\% accuracy for our test dataset regardless of the challenging nature of jersey number
identification in a low-data regime. This performance is on par with some of the most recent works~\citep{r7}. However,
the lack of a benchmark dataset for this task and unavailability of already implemented tools, is a big barrier
for comparing performance across all methods. The only solution is to label large amounts of high-quality data
and retrain the available solutions in-house. This requires a lot of computational resources and man-hours put
into work, which is not always an option for all institutions.

In our jersey detection models, we used ResNet50 as a base model, because it proved to be effective for this task.
Bigger and more sophisticated models might provide better accuracy and recall but an exhaustive search is necessary
for each of the components of the solutions to determine an optimal cost-benefit tradeoff. We recognize that more
investigation is needed here to determine such optimal.

In our solution we chose a three-model pipeline approach versus a one-pass prediction model. Our approach comes with
a few limitations including cascading inaccuracies from one model to the next and increase in latency. However, our
choice was justified by ease of implementation, maintenance and portability to other domains. Even with this cascading
effect, our solution proves to have a good performance in our highly imbalanced, limited dataset.

\section{Future Work}\label{sec:future-work}
Our approach to increase performance can be broadly classified into two categories: improving data quality and quantity
or experimenting with different models.

\subsection{Data quality and quantity}\label{subsec:data-quality-and-quantity}
We observed no improvement in model accuracy by increasing the number of duplicated samples or the number of image
augmentations. The confidence of the predictions directly correlated with the quality and resolution of the jersey
number crop (input image). In future work, we plan to experiment with various image quality enhancement methods in
classical CV and deep learning domains to observe if it improves performance. Another path that can be considered is
to refine our synthetic data generation pipeline to produce images that are closer to the real-world dataset.

\subsection{Different model strategies}\label{subsec:different-model-strategies}
Our current method has minimal labeling effort. However, by collecting more images of reasonable quality and quantity
we plan to test object detection-based models. One way to improve frame level accuracy would be to track detected
jersey numbers across both side-line and end-zone views so that in situations where numbers are partially visible
or player pose is complex, we would be able to obtain predictions with continuity. Tracking players in team sports
like football is still a major challenge in the sports CV domain and we will evaluate its utility in our future work.

\section{Conclusion}\label{sec:conclusion}
This paper presented a new solution for low-data regime jersey detection with two-stage novel synthetic data generation
techniques, pose estimation for jersey number localization and CNN ensemble learning to detect jersey numbers.
Data augmentations during training and the use of large synthetic dataset provided enough variations for the model
to generalize well and learn numbers. Our solution is easy to implement, requires minimal labeling, curation,
supervision, and can be customized for various sports jersey fonts, colors and backgrounds. Our framework improves
the accuracy of number detection task by 9\% and can be easily extended to similar tasks across various Sports
communities as well as industries with similar use cases. Furthermore, our solution did not require the modification
of the data capturing or processing pipeline that is already in place, making it convenient and flexible.

Additionally, it introduces a novel data synthesis technique that can boost custom solution performance in a wide
array of sports. We hope this solution enables the Sport Analytics community to rapidly automate video understanding solutions.



%

%

\vskip 0.2in
\bibliography{jersey-detection}

\begin{thebibliography}{28}
\providecommand{\natexlab}[1]{#1}
\providecommand{\url}[1]{\texttt{#1}}
\expandafter\ifx\csname urlstyle\endcsname\relax
  \providecommand{\doi}[1]{doi: #1}\else
  \providecommand{\doi}{doi: \begingroup \urlstyle{rm}\Url}\fi

\bibitem[Borkman et~al.(2021)Borkman, Crespi, Dhakad, Ganguly, Hogins, Jhang,
  Kamalzadeh, Li, Leal, Parisi, et~al.]{r20}
S.~Borkman, A.~Crespi, S.~Dhakad, S.~Ganguly, J.~Hogins, Y.-C. Jhang,
  M.~Kamalzadeh, B.~Li, S.~Leal, P.~Parisi, et~al.
\newblock Unity perception: Generate synthetic data for computer vision.
\newblock \emph{arXiv preprint arXiv:2107.04259}, 2021.

\bibitem[De~Campos et~al.(2009)De~Campos, Babu, Varma, et~al.]{r11}
T.~E. De~Campos, B.~R. Babu, M.~Varma, et~al.
\newblock Character recognition in natural images.
\newblock \emph{VISAPP (2)}, 7:\penalty0 2, 2009.

\bibitem[DeTone et~al.(2018)DeTone, Malisiewicz, and Rabinovich]{r13}
D.~DeTone, T.~Malisiewicz, and A.~Rabinovich.
\newblock Superpoint: Self-supervised interest point detection and description.
\newblock In \emph{Proceedings of the IEEE conference on computer vision and
  pattern recognition workshops}, pages 224--236, 2018.

\bibitem[Duan et~al.(2019)Duan, Bai, Xie, Qi, Huang, and Tian]{r23}
K.~Duan, S.~Bai, L.~Xie, H.~Qi, Q.~Huang, and Q.~Tian.
\newblock Centernet: Keypoint triplets for object detection.
\newblock In \emph{Proceedings of the IEEE/CVF international conference on
  computer vision}, pages 6569--6578, 2019.

\bibitem[Gerke et~al.(2015)Gerke, Muller, and Schafer]{r4}
S.~Gerke, K.~Muller, and R.~Schafer.
\newblock Soccer jersey number recognition using convolutional neural networks.
\newblock In \emph{Proceedings of the IEEE International Conference on Computer
  Vision Workshops}, pages 17--24, 2015.

\bibitem[Goodfellow et~al.(2013)Goodfellow, Bulatov, Ibarz, Arnoud, and
  Shet]{r25}
I.~J. Goodfellow, Y.~Bulatov, J.~Ibarz, S.~Arnoud, and V.~Shet.
\newblock Multi-digit number recognition from street view imagery using deep
  convolutional neural networks.
\newblock \emph{arXiv preprint arXiv:1312.6082}, 2013.

\bibitem[Hacohen and Weinshall(2019)]{r28}
G.~Hacohen and D.~Weinshall.
\newblock On the power of curriculum learning in training deep networks.
\newblock In \emph{International Conference on Machine Learning}, pages
  2535--2544. PMLR, 2019.

\bibitem[He et~al.(2016)He, Zhang, Ren, and Sun]{r29}
K.~He, X.~Zhang, S.~Ren, and J.~Sun.
\newblock Deep residual learning for image recognition.
\newblock In \emph{Proceedings of the IEEE conference on computer vision and
  pattern recognition}, pages 770--778, 2016.

\bibitem[Hinterstoisser et~al.(2018)Hinterstoisser, Lepetit, Wohlhart, and
  Konolige]{r19}
S.~Hinterstoisser, V.~Lepetit, P.~Wohlhart, and K.~Konolige.
\newblock On pre-trained image features and synthetic images for deep learning.
\newblock In \emph{Proceedings of the European Conference on Computer Vision
  (ECCV) Workshops}, pages 0--0, 2018.

\bibitem[Hinterstoisser et~al.(2019)Hinterstoisser, Pauly, Heibel, Martina, and
  Bokeloh]{r18}
S.~Hinterstoisser, O.~Pauly, H.~Heibel, M.~Martina, and M.~Bokeloh.
\newblock An annotation saved is an annotation earned: Using fully synthetic
  training for object detection.
\newblock In \emph{Proceedings of the IEEE/CVF international conference on
  computer vision workshops}, pages 0--0, 2019.

\bibitem[Jeon et~al.(2021)Jeon, Kim, and Kim]{r21}
E.~Jeon, K.~Kim, and D.~Kim.
\newblock Fa-gan: Feature-aware gan for text to image synthesis.
\newblock In \emph{2021 IEEE International Conference on Image Processing
  (ICIP)}, pages 2443--2447. IEEE, 2021.

\bibitem[Larochelle et~al.(2008)Larochelle, Erhan, and Bengio]{r10}
H.~Larochelle, D.~Erhan, and Y.~Bengio.
\newblock Zero-data learning of new tasks.
\newblock In \emph{AAAI}, volume~1, page~3, 2008.

\bibitem[Li et~al.(2018)Li, Xu, Liu, Li, and Wang]{r5}
G.~Li, S.~Xu, X.~Liu, L.~Li, and C.~Wang.
\newblock Jersey number recognition with semi-supervised spatial transformer
  network.
\newblock In \emph{Proceedings of the IEEE Conference on Computer Vision and
  Pattern Recognition Workshops}, pages 1783--1790, 2018.

\bibitem[Lin et~al.(2014)Lin, Maire, Belongie, Hays, Perona, Ramanan,
  Doll{\'a}r, and Zitnick]{r8}
T.-Y. Lin, M.~Maire, S.~Belongie, J.~Hays, P.~Perona, D.~Ramanan,
  P.~Doll{\'a}r, and C.~L. Zitnick.
\newblock Microsoft coco: Common objects in context.
\newblock In \emph{European conference on computer vision}, pages 740--755.
  Springer, 2014.

\bibitem[Liu and Bhanu(2019)]{r6}
H.~Liu and B.~Bhanu.
\newblock Pose-guided r-cnn for jersey number recognition in sports.
\newblock In \emph{Proceedings of the IEEE/CVF Conference on Computer Vision
  and Pattern Recognition Workshops}, pages 0--0, 2019.

\bibitem[Lu et~al.(2013)Lu, Lin, Hsu, Weng, Kang, and Liao]{r3}
C.-W. Lu, C.-Y. Lin, C.-Y. Hsu, M.-F. Weng, L.-W. Kang, and H.-Y.~M. Liao.
\newblock Identification and tracking of players in sport videos.
\newblock In \emph{Proceedings of the Fifth International Conference on
  Internet Multimedia Computing and Service}, pages 113--116, 2013.

\bibitem[Mustikovela et~al.(2021)Mustikovela, De~Mello, Prakash, Iqbal, Liu,
  Nguyen-Phuoc, Rother, and Kautz]{r22}
S.~K. Mustikovela, S.~De~Mello, A.~Prakash, U.~Iqbal, S.~Liu, T.~Nguyen-Phuoc,
  C.~Rother, and J.~Kautz.
\newblock Self-supervised object detection via generative image synthesis.
\newblock In \emph{Proceedings of the IEEE/CVF International Conference on
  Computer Vision}, pages 8609--8618, 2021.

\bibitem[Nikolenko(2021)]{r15}
S.~I. Nikolenko.
\newblock Synthetic simulated environments.
\newblock In \emph{Synthetic Data for Deep Learning}, pages 195--215. Springer,
  2021.

\bibitem[Nikolenko et~al.(2021)]{r12}
S.~I. Nikolenko et~al.
\newblock \emph{Synthetic data for deep learning}.
\newblock Springer, 2021.

\bibitem[Piacentino et~al.(2021)Piacentino, Guarner, and Angulo]{r17}
E.~Piacentino, A.~Guarner, and C.~Angulo.
\newblock Generating synthetic ecgs using gans for anonymizing healthcare data.
\newblock \emph{Electronics}, 10\penalty0 (4):\penalty0 389, 2021.

\bibitem[Roberts et~al.(2021)Roberts, Ramapuram, Ranjan, Kumar, Bautista,
  Paczan, Webb, and Susskind]{r16}
M.~Roberts, J.~Ramapuram, A.~Ranjan, A.~Kumar, M.~A. Bautista, N.~Paczan,
  R.~Webb, and J.~M. Susskind.
\newblock Hypersim: A photorealistic synthetic dataset for holistic indoor
  scene understanding.
\newblock In \emph{Proceedings of the IEEE/CVF International Conference on
  Computer Vision}, pages 10912--10922, 2021.

\bibitem[{\v{S}}ari et~al.(2008){\v{S}}ari, Dujmi, Papi, and Ro{\v{z}}i]{r2}
M.~{\v{S}}ari, H.~Dujmi, V.~Papi, and N.~Ro{\v{z}}i.
\newblock Player number localization and recognition in soccer video using hsv
  color space and internal contours.
\newblock In \emph{The International Conference on Signal and Image Processing
  (ICSIP 2008)}. Citeseer, 2008.

\bibitem[Settles(2009)]{r9}
B.~Settles.
\newblock Active learning literature survey.
\newblock 2009.

\bibitem[Siam et~al.(2021)Siam, Kendall, and Jagersand]{r14}
M.~Siam, A.~Kendall, and M.~Jagersand.
\newblock Video class agnostic segmentation benchmark for autonomous driving.
\newblock In \emph{Proceedings of the IEEE/CVF Conference on Computer Vision
  and Pattern Recognition}, pages 2825--2834, 2021.

\bibitem[Sun(2019)]{r26}
S.~Sun.
\newblock Multi-digit mnist for few-shot learning.
\newblock \emph{URL: https://github. com/shaohua0116/MultiDigitMNIST}, 2019.

\bibitem[Vats et~al.(2021)Vats, Fani, Clausi, and Zelek]{r7}
K.~Vats, M.~Fani, D.~A. Clausi, and J.~Zelek.
\newblock Multi-task learning for jersey number recognition in ice hockey.
\newblock In \emph{Proceedings of the 4th International Workshop on Multimedia
  Content Analysis in Sports}, pages 11--15, 2021.

\bibitem[Weinshall et~al.(2018)Weinshall, Cohen, and Amir]{r27}
D.~Weinshall, G.~Cohen, and D.~Amir.
\newblock Curriculum learning by transfer learning: Theory and experiments with
  deep networks.
\newblock In \emph{International Conference on Machine Learning}, pages
  5238--5246. PMLR, 2018.

\bibitem[Ye et~al.(2005)Ye, Huang, Jiang, Liu, and Gao]{r1}
Q.~Ye, Q.~Huang, S.~Jiang, Y.~Liu, and W.~Gao.
\newblock Jersey number detection in sports video for athlete identification.
\newblock In \emph{Visual Communications and Image Processing 2005}, volume
  5960, pages 1599--1606. SPIE, 2005.

\end{thebibliography}

\end{document}